\documentclass[runningheads]{llncs}
\usepackage[T1]{fontenc}

\usepackage[utf8]{inputenc} 
\usepackage[T1]{fontenc}    
\usepackage{url}            
\usepackage{booktabs}       
\usepackage{amsfonts}       
\usepackage{amsmath}
\usepackage{nicefrac}       
\usepackage{microtype}      
\usepackage{xcolor}         
\usepackage{array}
\usepackage{graphicx}
\usepackage{subcaption}
\usepackage{tikz}
\usepackage{pgfplots}
\usepackage{multirow}
\usepackage{comment}

\usepackage[pagebackref,breaklinks,colorlinks]{hyperref}

\usepackage[capitalize]{cleveref}
\crefname{section}{Sec.}{Secs.}
\Crefname{section}{Section}{Sections}
\Crefname{table}{Table}{Tables}
\crefname{table}{Tab.}{Tabs.}

\pgfdeclareplotmark{mystar}{
    \node[star,star point ratio=2.25,
          inner sep=0pt] {};
}

\newcolumntype{L}[1]{>{\raggedright\let\newline\\\arraybackslash\hspace{0pt}}m{#1}}
\newcolumntype{C}[1]{>{\centering\let\newline\\\arraybackslash\hspace{0pt}}m{#1}}
\newcolumntype{R}[1]{>{\raggedleft\let\newline\\\arraybackslash\hspace{0pt}}m{#1}}

\definecolor{lightred}{rgb}{1, 0.66, 0.66}

\begin{document}

\title{Aligning benchmark datasets for table structure recognition}
\titlerunning{Aligning benchmark datasets for TSR}

\author{Brandon Smock\inst{1}\orcidID{0009-0002-7002-0800} \and
Rohith Pesala\inst{1}\orcidID{0009-0004-7373-853X} \and
Robin Abraham\inst{1}\orcidID{0000-0003-1915-8118}}

\authorrunning{B. Smock et al.}

\institute{Microsoft, Redmond WA, USA\\
\email{\{brsmock,ropesala,robin.abraham\}@microsoft.com}}

\maketitle

\begin{abstract}
Benchmark datasets for table structure recognition (TSR) must be carefully processed to ensure they are annotated consistently.
However, even if a dataset's annotations are self-consistent, there may be significant inconsistency across datasets, which can harm the performance of models trained and evaluated on them.
In this work, we show that \emph{aligning} these benchmarks---removing both errors and inconsistency between them---improves model performance significantly.
We demonstrate this through a data-centric approach where we adopt one model architecture, the Table Transformer (TATR), that we hold fixed throughout.
Baseline exact match accuracy for TATR evaluated on the ICDAR-2013 benchmark is 65\% when trained on PubTables-1M, 42\% when trained on FinTabNet, and 69\% combined.
After reducing annotation mistakes and inter-dataset inconsistency, performance of TATR evaluated on ICDAR-2013 increases substantially to 75\% when trained on PubTables-1M, 65\% when trained on FinTabNet, and 81\% combined.
We show through ablations over the modification steps that canonicalization of the table annotations has a significantly positive effect on performance, while other choices balance necessary trade-offs that arise when deciding a benchmark dataset's final composition.
Overall we believe our work has significant implications for benchmark design for TSR and potentially other tasks as well.
Dataset processing and training code will be released at \url{https://github.com/microsoft/table-transformer}.
\end{abstract}

\section{Introduction}

Table extraction (TE) is a long-standing problem in document intelligence.
Over the last decade, steady progress has been made formalizing TE as a machine learning (ML) task.
This includes the development of task-specific metrics for evaluating table structure recognition (TSR) models~\cite{gobel2013icdar,zhong2019image,smock2022grits} as well as the increasing variety of datasets and benchmarks~\cite{chi2019complicated,zheng2021global,zhong2019image,smock2022pubtables1m}.
These developments have enabled significant advances in deep learning (DL) modeling for TE~\cite{schreiber2017deepdesrt,prasad2020cascadetabnet,zheng2021global,smock2022pubtables1m,Nassar_2022_CVPR,Liu_2022_CVPR}.

In general, benchmarks play a significant role in shaping the direction of ML research~\cite{raji2021ai,koch2021reduced}.
Recently it has been shown that benchmark datasets for TSR contain a significant number of errors and inconsistencies~\cite{smock2022pubtables1m}.
It is well-documented that errors in a benchmark have negative consequences for both learning and evaluation~\cite{guyon1994discovering,zhu2004class,frenay2013classification}.
Errors in the test set for the ImageNet benchmark~\cite{northcutt2021pervasive}, for instance, have led to top-1 accuracy saturating around 91\%~\cite{yu2022coca}.
Errors can also lead to false conclusions during model selection, particularly when the training data is drawn from the same noisy distribution as the test set~\cite{northcutt2021pervasive}.

Compared to annotation mistakes, inconsistencies in a dataset can be more subtle because they happen across a collection of samples rather than in isolated examples---but no less harmful.
Even if a single dataset is self-consistent there may be inconsistencies in labeling across different datasets for the same task.
We consider datasets for the same task that are annotated inconsistently with respect to each other to be \emph{misaligned}.
Misalignment can be considered an additional source of labeling noise.
This noise may go unnoticed when a dataset is studied in isolation, but it can have significant effects on model performance when datasets are combined.

\begin{figure*}[t]
    \centering
    \begin{tikzpicture}
    \definecolor{light gray}{RGB}{220, 220, 220}
    	\begin{axis} [grid, grid style={light gray, line cap=round}, ybar,enlarge x limits={abs=0.65cm},x=1.16cm,bar width=0.22cm,height=5.5cm,width=4cm,ylabel={$\text{Acc}_\text{Con}$},symbolic x coords={FinTabNet.a1, FinTabNet.a2, FinTabNet.a3, FinTabNet.a4, FinTabNet.a5, FinTabNet.a6, PubTables-1M, PubTables-1M + FinTabNet.a1, PubTables-1M + FinTabNet.a6},xtick = data, ymin=0.36, ymax=0.84,
    	xticklabel style={rotate=52, xshift=-0.4cm,yshift=0.75cm, align=right, text width=2.5cm},
    	legend pos=north west,
    	ytick={0.40, 0.44, 0.48, 0.52, 0.56, 0.60, 0.64, 0.68, 0.72, 0.76, 0.80},
    	tick label style={
                /pgf/number format/fixed,
                /pgf/number format/precision=3
            }]
    	\addplot coordinates {
    	    (FinTabNet.a1, 0.417)
                    (FinTabNet.a2, 0.436)
                    (FinTabNet.a3, 0.462)
                    (FinTabNet.a4, 0.551)
                    (FinTabNet.a5, 0.494)
                    (FinTabNet.a6, 0.500)
                    (PubTables-1M, 0.647)
                    (PubTables-1M + FinTabNet.a1, 0.686)
                    (PubTables-1M + FinTabNet.a6, 0.679)
    	};
    	\addplot coordinates {
    	    (FinTabNet.a1, 0.411)
                    (FinTabNet.a2, 0.430)
                    (FinTabNet.a3, 0.456)
                    (FinTabNet.a4, 0.544)
                    (FinTabNet.a5, 0.487)
                    (FinTabNet.a6, 0.506)
                    (PubTables-1M, 0.639)
                    (PubTables-1M + FinTabNet.a1, 0.677)
                    (PubTables-1M + FinTabNet.a6, 0.671)
    	};
    	\addplot coordinates {
    	    (FinTabNet.a1, 0.411)
                    (FinTabNet.a2, 0.405)
                    (FinTabNet.a3, 0.418)
                    (FinTabNet.a4, 0.589)
                    (FinTabNet.a5, 0.551)
                    (FinTabNet.a6, 0.646)
                    (PubTables-1M, 0.753)
                    (PubTables-1M + FinTabNet.a1, 0.785)
                    (PubTables-1M + FinTabNet.a6, 0.810)
    	};
    	\legend {ICDAR2013.a1, ICDAR2013.a2, ICDAR2013.a3};
    	\end{axis}
    \end{tikzpicture}
    \caption{The improvement in performance (in terms of exact match accuracy) of models trained on FinTabNet and PubTables-1M, and evaluated on ICDAR-2013, as we remove annotation mistakes and inconsistencies between the three datasets.}
    \label{fig:summary}
\end{figure*}

In this work we study the effect that errors and misalignment between benchmark datasets have on model performance for TSR.
We select two large-scale crowd-sourced datasets for training---FinTabNet and PubTables-1M---and one small expert-labeled dataset for evaluation---the ICDAR-2013 benchmark.
For our models we choose a single fixed architecture, the recently proposed Table Transformer (TATR)~\cite{smock2022pubtables1m}.
This can be seen as a data-centric~\cite{strickland2022andrew} approach to ML, where we hold the modeling approach fixed and instead seek to improve performance through data improvements.
Among our main contributions:
\begin{itemize}
    \item We remove both annotation mistakes and inconsistencies between FinTabNet and ICDAR-2013, aligning these datasets with PubTables-1M and producing improved versions of two standard benchmark datasets for TSR (which we refer to as FinTabNet.c and ICDAR-2013.c, respectively).
    \item We show that removing inconsistencies between benchmark datasets for TSR indeed has a substantial positive impact on model performance, improving baseline models trained separately on PubTables-1M and FinTabNet from 65\% to 78\% and 42\% to 65\%, respectively, when evaluated on ICDAR-2013 (see \cref{fig:summary}).
    \item  We perform a sequence of ablations over the steps of the correction procedure, which shows that canonicalization has a clear positive effect on model performance, while other factors create trade-offs when deciding the final composition of a benchmark.
    \item We train a single model on both PubTables-1M and the aligned version of FinTabNet (FinTabNet.c), establishing a new baseline DAR of 0.965 and exact match table recognition accuracy of 81\% on the corrected ICDAR-2013 benchmark (ICDAR-2013.c).
    \item We plan to release all of the dataset processing and training code at \url{https://github.com/microsoft/table-transformer}.
\end{itemize}

\begin{figure*}[t]
  \centering
  \begin{subfigure}[b]{0.8\linewidth}
	\centering
	\includegraphics[width=7cm]{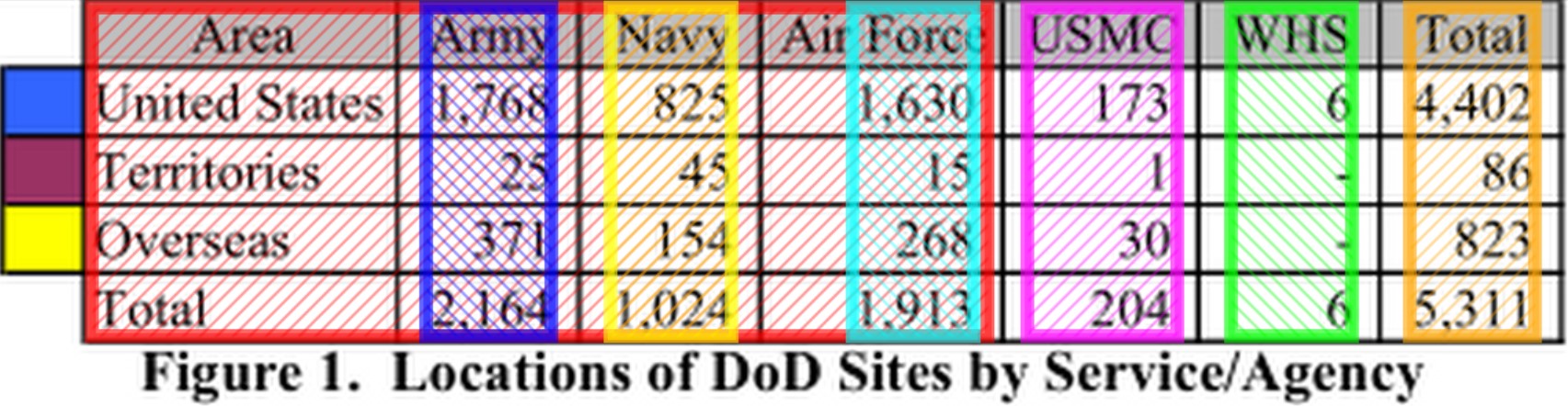}
    \caption{Original annotation}
    \label{subfig:overlap}
  \end{subfigure}
  \begin{subfigure}[b]{0.8\linewidth}
	\centering
	\includegraphics[width=7cm]{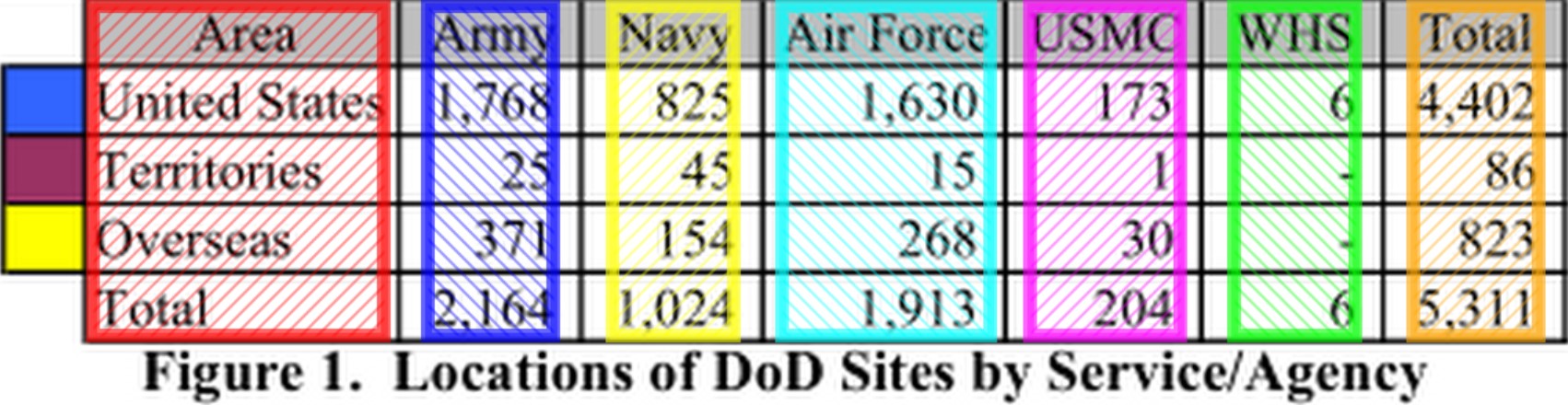}
    \caption{Corrected annotation}
    \label{subfig:without_overlap}
  \end{subfigure}
  \caption{An example table from us-gov-dataset/us-012-str.xml in the ICDAR-2013 dataset. On the top are the column bounding boxes created when using the dataset's original annotations. On the bottom are the column bounding boxes following correction of the cells' column index labels. Corresponding columns before and after correction are shaded with the same color.}
  \label{fig:icdar_error}
\end{figure*}

\section{Related Work}

The first standard multi-domain benchmark dataset for table structure recognition was the ICDAR-2013 dataset, introduced at the 2013 ICDAR Table Competition \cite{gobel2013icdar}.
In total it contains 158 tables\footnote{The dataset is originally documented as having 156 tables but we found 2 cases of tables that need to be split into multiple tables to be annotated consistently.} in the official competition dataset and at least 100 tables\footnote{This is the number of tables in the practice set we were able to find online.} in a practice dataset.
The competition also established the first widely-used metric for evaluation, the directed adjacency relations (DAR) metric, though alternative metrics such as TEDS~\cite{zhong2019image} and GriTS \cite{smock2022grits} have been proposed more recently.
At the time of the competition, the best reported approach achieved a DAR of 0.946.
More recent work from Tensmeyer et al. \cite{tensmeyer2019deep} reports a DAR of 0.953 on the full competition dataset.

The difficulty of the ICDAR-2013 benchmark and historical lack of training data for the TSR task has led to a variety of approaches and evaluation procedures that differ from those established in the original competition.
DeepDeSRT \cite{schreiber2017deepdesrt} split the competition dataset into a training set and a test set of just 34 samples.
Additionally, the authors processed the annotations to add row and column bounding boxes and then framed the table structure recognition problem as the task of detecting these boxes.
This was a significant step at the time but because spanning cells are ignored it is only a partial solution to the full recognition problem.
Still, many works have followed this approach~\cite{hashmi2021current}.
Hashmi et al.~\cite{hashmi2021guided} recently reported an F-score of 0.9546 on the row and column detection task, compared to the original score of 0.9144 achieved by DeepDeSRT.

While ICDAR-2013 has been a standard benchmark of progress over the last decade, little attention has been given to complete solutions to the TSR task evaluated on the full ICDAR-2013 dataset.
Of these, none reports exact match recognition accuracy.
Further, we are not aware of any work that points out label noise in the ICDAR-2013 dataset and attempts to correct it.
Therefore it is an open question to what extent performance via model improvement has saturated on this benchmark in its current form.

\begin{figure*}[t]
  \centering
  \includegraphics[width=11cm]{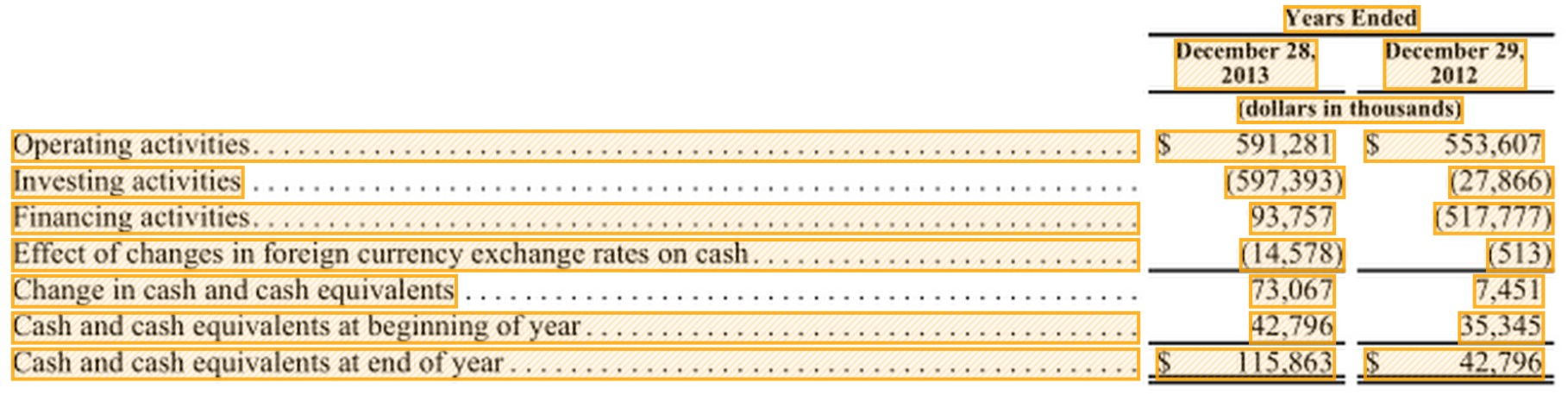}
  \caption{An example table from HBI/2013/page\textunderscore 39.pdf in the FinTabNet dataset with cell bounding boxes (shaded in orange) that are inconsistently annotated. Some cells in the first column include the dot leaders in the bounding box while others do not.}
  \label{fig:fintabnet_1}
\end{figure*}

To address the need for training data, several large-scale crowd-sourced datasets \cite{chi2019complicated,zhong2019image,li2020tablebank,zheng2021global,smock2022pubtables1m} have been released for training table structure recognition models.
However, the quality of these datasets is lower overall than that of expert-labeled datasets like ICDAR-2013.
Of particular concern is the potential for ambiguity in table structures \cite{hu2001table,seth2010analysis,broman2018data,paramonov2020table}.
A specific form of label error called oversegmentation \cite{smock2022pubtables1m} is widely prevalent in benchmark TSR datasets, which ultimately leads to annotation inconsistency.
However, models developed and evaluated on these datasets typically treat them as if they are annotated unambiguously, with only one possible correct interpretation.

To address this, Smock et al. \cite{smock2022pubtables1m} proposed a canonicalization algorithm, which can automatically correct and make table annotations more consistent.
However, this was applied to only a single source of data.
The impact of errors and inconsistency across multiple benchmark datasets remains an open question.
In fact, the full extent of this issue can be masked by using metrics that measure the average correctness of cells \cite{smock2022grits}.
But for table extraction systems in real-world settings, individual errors may not be tolerable, and therefore the accuracy both of cells and of entire tables is important to consider.

\section{Data}

As our baseline training datasets we adopt PubTables-1M \cite{smock2022pubtables1m} and FinTabNet \cite{zheng2021global}.
PubTables-1M contains nearly one million tables from scientific articles, while FinTabNet contains close to 113k tables from financial reports.
As our baseline evaluation dataset we adopt the ICDAR-2013 \cite{gobel2013icdar} dataset.
The ICDAR-2013 dataset contains tables from multiple document domains manually annotated by experts.
While its size limits its usefulness for training, the quality of ICDAR-2013 and its diverse set of table appearances and layouts make it useful for model benchmarking.
For the ICDAR-2013 dataset, we use the competition dataset as the test set, use the practice dataset as a validation set, and consider each table "region" annotated in the dataset to be a table, yielding a total of 256 tables to start.

\begin{table*}[!t]
  \scriptsize
  \caption{The high-level data processing steps and ablations for the FinTabNet and ICDAR-2013 datasets. Data cleaning steps are grouped into stages. Each stage represents an addition to the steps of the full processing algorithm. For each stage, a dataset ablation is created with the steps up to and including that stage. But the stages themselves do not occur in the same exact order in the full processing pipeline. For instance, when stage a5 is added to the processing it actually occurs at the beginning of stage a4 to annotate more column headers prior to canonicalization.}
  \label{tab:corrections}
  \centering
  \begin{tabular}[t]{L{1.6cm}L{2.2cm}p{8cm}}
    \toprule
    \textbf{Dataset} & \textbf{Stage} & \textbf{Description} \\
    \midrule
    FinTabNet & & \\
    \hspace{2mm}--- & Unprocessed & The original data, which does not have header or row/column bounding box annotations. \\
     \hspace{2mm}a1 & Completion & Baseline FinTabNet dataset; create bounding boxes for all rows and columns; remove tables for which these boxes cannot be defined.\\
     \hspace{2mm}a2 & \multirow[t]{4}{1.7cm}{Cell box adjustment} & \multirow[t]{4}{8cm}{Iteratively refine the row and column boxes to tightly surround their coinciding text; create \emph{grid cells} at the intersection of each row and column; remove any table for which 50\% of the area of a word overlaps with multiple cells in the table.}\\
     & &\\
     & &\\
     & &\\
     \hspace{2mm}a3 & \multirow[t]{5}{1.7cm}{Consistency adjustments} & \multirow[t]{5}{8cm}{Make the sizing and spanning of columns and rows more consistent; adjust cell bounding boxes to always ignore dot leaders (remove any tables for which this was unsuccessful); remove empty rows and empty columns from the annotations; merge any adjacent header rows whose cells all span the same columns; remove tables with columns containing just cent and dollar signs (we found these difficult to automatically correct).}\\
     & &\\
     & &\\
     & &\\
     & &\\
     & &\\
     & &\\
     \hspace{2mm}a4 & \multirow[t]{4}{2cm}{Canonicalization} & \multirow[t]{4}{8cm}{Use cell structure to infer the column header and projected row headers for each table; remove tables with only two columns as the column header of these is structurally ambiguous; canonicalize each annotation.} \\
     & &\\
     & &\\
     & &\\
     \hspace{2mm}a5 & \multirow[t]{3}{2cm}{Additional column header inference} & \multirow[t]{3}{8cm}{Infer the column header for many tables with only two columns by using the cell text in the first row to determine if the first row is a column header.}\\
     & &\\
     & &\\
     \hspace{2mm}a6 & Quality control & Remove tables with erroneous annotations, including: tables where words in the table coincide with either zero or more than one cell bounding box, tables with a projected row header at the top or bottom (indicating a caption or footer is mistakenly included in the table), and tables that appear to have only a header and no body.\\
    \midrule
    ICDAR-2013 & &\\
    \hspace{2mm}--- & Unprocessed & The original data, which does not have header or row/column bounding box annotations. \\
     \hspace{2mm}a1 & Completion & Baseline ICDAR-2013 dataset; create bounding boxes for all rows and columns; remove tables for which these boxes cannot be defined or whose annotations cannot be processed due to errors.\\
     \hspace{2mm}a2 & Manual correction & Manually fix annotation errors (18 tables fixed in total).\\
     \hspace{2mm}a3 &\multirow[t]{3}{2cm}{Consistency adjustments and canonicalization} & \multirow[t]{3}{8cm}{Apply the same automated steps applied to FinTabNet.a2 through FinTabNet.a4; after processing, manually inspect each table and make any additional corrections as needed.}\\
     & &\\
     & &\\
    \midrule
  \end{tabular}
\end{table*}

\subsection{Missing Annotations}
While the annotations in FinTabNet and ICDAR-2013 are sufficient to train and evaluate models for table structure recognition, several kinds of labels that could be of additional use are not explicitly annotated.
Both datasets annotate bounding boxes for each cell as the smallest rectangle fully-enclosing all of the text of the cell.
However, neither dataset annotates bounding boxes for rows, columns, or blank cells.
Similarly, neither dataset explicitly annotates which cells belong to the row and column headers.

For each of these datasets, one of the first steps we take before correcting their labels is to automatically add labels that are missing.
This can be viewed as making explicit the labels that are \emph{implicit} given the rest of the label information present.
We cover these steps in more detail later in this section.

\begin{table*}[!t]
  \scriptsize
  \caption{Diversity and complexity of table instances in the baseline and ablation datasets.}
  \label{tab:stats}
  \centering
  \begin{tabular}{L{0.17\textwidth}R{0.13\textwidth}R{0.14\textwidth}R{0.12\textwidth}R{0.11\textwidth}R{0.1\textwidth}R{0.16\textwidth}}
    \toprule
    \textbf{Dataset} & \textbf{\# Tables$^\ddag$} & \textbf{\# Unique \mbox{Topologies}} & \textbf{Avg. Tables / \mbox{Topology}} & \textbf{Avg. Rows / \mbox{Table}} & \textbf{Avg. Cols. / \mbox{Table}} & \textbf{Avg. Spanning Cells / \mbox{Table}} \\
    \midrule
    FinTabNet & \textbf{112,875} & 9,627 & 11.72 & 11.94 & 4.36 & 1.01 \\
    FinTabNet.a1 & 112,474 & 9,387 & 11.98 & 11.92 & 4.35 & 1.00 \\
    FinTabNet.a2 & 109,367 & 8,789 & 12.44 & 11.87 & 4.33 & 0.98 \\
    FinTabNet.a3 & 103,737 & 7,647 & 13.57 & 11.81 & 4.28 & 0.93 \\
    FinTabNet.a4 & 89,825 & 18,480 & \textbf{4.86} & \textbf{11.97} & \textbf{4.61} & \textbf{2.79} \\
    FinTabNet.a5 & 98,019 & \textbf{18,752} & 5.23 & 11.78 & 4.39 & 2.57 \\
    FinTabNet.a6 & 97,475 & 18,702 & 5.21 & 11.81 & 4.39 & 2.58 \\
    \midrule
    ICDAR-2013 & 256 & 181 & 1.41 & \textbf{15.88} & \textbf{5.57} & 1.36 \\
    ICDAR-2013.a1 & 247 & 175 & 1.41 & 15.61 & 5.39 & 1.34 \\
    ICDAR-2013.a2 & \textbf{258} & 181 & 1.43 & 15.55 & 5.45 & 1.42 \\
    ICDAR-2013.a3 & \textbf{258} & \textbf{184} & \textbf{1.40} & 15.52 & 5.45 & \textbf{2.08} \\
    \midrule
    \multicolumn{7}{l}{\scriptsize $^\ddag$The number of tables in the dataset that we were able to successfully read and process.} \\
  \end{tabular}
\end{table*}

\subsection{Label Errors and Inconsistencies}
Next we investigate both of these datasets to look for possible annotation mistakes and inconsistencies.
For FinTabNet, due to the impracticality of verifying each annotation manually, we initially sample the dataset to identify types of errors common to many instances.
For ICDAR-2013, we manually inspect all 256 table annotations.

For both datasets, we note that the previous action of defining and adding bounding boxes for rows and columns was crucial to catching inconsistencies, as it exposed a number of errors when we visualized these boxes on top of the tables.
For ICDAR-2013 we noticed the following errors during manual inspection:
14 tables with at least one cell with incorrect column or row indices (see \cref{fig:icdar_error}), 4 tables with at least one cell with an incorrect bounding box, 1 table with a cell with incorrect text content, and 2 tables that needed to be split into more than one table.

For FinTabNet, we noticed a few errors that appear to be common in crowd-sourced table annotations in the financial domain.
For example, we noticed many of these tables use \emph{dot leaders} for spacing and alignment purposes.
However, we noticed a significant amount of inconsistency in whether dot leaders are included or excluded from a cell's bounding box (see \cref{fig:fintabnet_1}).
We also noticed that it is common in these tables for a logical column that contains a monetary symbol, such as a \$, to be split into two columns, possibly for numerical alignment purposes (for an example, see NEE/2006/page\textunderscore 49.pdf).
Tables annotated this way are done so for presentation purposes but are inconsistent with their logical interpretation, which should group the dollar sign and numerical value into the same cell.
Finally we noticed annotations with fully empty rows.
Empty rows can sometimes be used in a table's markup to create more visual spacing between rows.
But there is no universal convention for how much spacing corresponds to an empty row.
Furthermore, empty rows used only for spacing serve no logical purpose.
Therefore, empty rows in these tables should be considered labeling errors and should be removed for consistency.
For both datasets we noticed oversegmentation of header cells, as is common in crowd-sourced markup tables.

\subsection{Dataset Corrections and Alignment}

Mistakes noticed during manual inspection of ICDAR-2013 are corrected directly.
For both datasets, nearly all of the other corrections and alignments are done by a series of automated processing steps.
We list these steps at a high level in \cref{tab:corrections}.
For each dataset the processing steps are grouped into a series of macro-steps, which helps us to study some of their effects in detail using ablation experiments in \cref{sec:experiments}.

As part of the processing we adopt the canonicalization procedure \cite{smock2022pubtables1m} used to create PubTables-1M.
This helps to align all three datasets and minimize the amount of inconsistencies between them.
We make small improvements to the canonicalization procedure that help to generalize it to tables in domains other than scientific articles.
These include small improvements to the step that determines the column header and portions of the row header.
Tables with inconsistencies that are determined to not be easily or reliably corrected using the previous steps are filtered out.

When creating a benchmark dataset there are many objectives to consider such as maximizing the number of samples, the diversity of the samples, the accuracy and consistency of the labels, the richness of the labels, and the alignment between the labels and the desired use of the learned model for a task.
When cleaning a pre-existing dataset these goals potentially compete and trade-offs must be made.
There is also the added constraint in this case that whatever steps were used to create the pre-existing dataset prior to cleaning are unavailable to us and unchangeable.
This prevents us from optimizing the entire set of processing steps holistically.
Overall we balance the competing objectives under this constraint by correcting and adding to the labels as much as possible when we believe this can be done reliably and filtering out potentially low-quality samples in cases where we do not believe we can reliably amend them.
We document the effects that the processing steps have on the size, diversity, and complexity of the samples in \cref{tab:stats}.

\begin{table*}[t]
  \scriptsize
  \caption{Comparison of the original baseline model for PubTables-1M and the one we study in the current work, which uses additional image cropping during training and a slightly longer training schedule.}
  \label{tab:pubtables1m}
  \centering
  \begin{tabular}{L{0.17\textwidth}L{0.11\textwidth}L{0.1\textwidth}L{0.09\textwidth}R{0.12\textwidth}R{0.12\textwidth}R{0.12\textwidth}R{0.1\textwidth}}
    \toprule
    \textbf{Training Data} & \textbf{Baseline Version} & \textbf{Epoch$^\ddag$} & \textbf{Test Images} & $\textbf{GriTS}_\textbf{Con}$ & $\textbf{GriTS}_\textbf{Loc}$ & $\textbf{GriTS}_\textbf{Top}$ & $\textbf{Acc}_\textbf{Con}$\\
    \midrule
    PubTables-1M & Original & 23.72 & Padded & 0.9846 & 0.9781 & 0.9845 & 0.8138\\
     & Current & 29 & Tight & \textbf{0.9855} & \textbf{0.9797} & \textbf{0.9851} & \textbf{0.8326}\\
    \midrule
    \multicolumn{8}{l}{\scriptsize $^\ddag$In the current work an epoch is standardized across datasets to equal 720,000 training samples.} \\
  \end{tabular}
\end{table*}

\section{Experiments}\label{sec:experiments}

For our experiments, we adopt a data-centric approach using the Table Transformer (TATR)~\cite{smock2022pubtables1m}.
TATR frames TSR as object detection using six classes and is implemented with DETR \cite{carion2020end}.
We hold the TATR model architecture fixed and make changes only to the data.
TATR is free of any TSR-specific engineering or inductive biases, which importantly forces the model to learn to solve the TSR task from its training data alone.

We make a few slight changes to the original approach used to train TATR on PubTables-1M in order to establish a stronger baseline and standardize the approach across datasets.
First we re-define an epoch as 720,000 training samples, which corresponds to 23.72 epochs in the original paper, and extend the training to 30 epochs given the new definition.
Second, we increase the amount of cropping augmentation during training.
All of the table images in the PubTables-1M dataset contain additional padding around the table from the page that each table is cropped from.
This extra padding enable models to be trained with cropping augmentation without removing parts of the image belonging to the table.
However, the original paper evaluated models on the padded images rather than tightly-cropped table images.
In this paper, we instead evaluate all models on more tightly-cropped table images, leaving only 2 pixels of padding around the table boundary in each table image's original size.
The use of tight cropping better reflects how TSR models are expected to be used in real-world settings when paired with an accurate table detector.

\begin{figure*}[t]
    \centering
    \begin{tikzpicture}
    \definecolor{light gray}{RGB}{220, 220, 220}
    	\begin{axis} [grid, grid style={light gray, line cap=round}, ybar,enlarge x limits={abs=0.65cm},x=1cm,bar width=0.5cm,height=4cm,ylabel={$\text{Acc}_\text{Con}$},symbolic x coords={FinTabNet.a1,FinTabNet.a2,FinTabNet.a3,FinTabNet.a4,FinTabNet.a5,FinTabNet.a6},xtick = data, ymin=0.758, ymax=0.835,
    	xticklabel style={rotate=25, xshift=-0.8cm,yshift=0.45cm, align=right, text width=2.5cm},
    	legend pos=north west,
    	ytick={0.77, 0.79, 0.81, 0.83},
    	tick label style={
                /pgf/number format/fixed,
                /pgf/number format/precision=3
            }]
    	\addplot [draw=red, fill=lightred] coordinates {
    	    (FinTabNet.a1, 0.787)
                    (FinTabNet.a2, 0.768)
                    (FinTabNet.a3, 0.799)
                    (FinTabNet.a4, 0.793)
                    (FinTabNet.a5, 0.828)
                    (FinTabNet.a6, 0.812)
    	};
    	\end{axis}
    \end{tikzpicture}
    \caption{The performance of TATR trained and evaluated on each version of FinTabNet. As can be seen, performance of FinTabNet evaluated on itself generally improves as a result of the modification steps. Performance is highest for FinTabNet.a5 rather than for the final version of FinTabNet. However, FinTabNet.a6 ends up being better aligned with other benchmark datasets, highlighting the potential trade-offs that must be considering when designing a benchmark dataset.}
    \label{fig:fin_summary}
\end{figure*}

For evaluation, we use the recently proposed grid table similarity (GriTS) \cite{smock2022grits} metrics as well as table content exact match accuracy ($\textrm{Acc}_\textrm{Con}$).
GriTS compares predicted tables and ground truth directly in matrix form and can be interpreted as an F-score over the correctness of predicted cells.
Exact match accuracy considers the percentage of tables for which all cells, including blank cells, are matched exactly.
The TATR model requires words and their bounding boxes to be extracted separately and uses maximum overlap between words and predicted cells to slot the words into their cells.
For evaluation, we assume that the true bounding boxes for words are given to the model and used to determine the final output.

In \cref{tab:pubtables1m} we evaluate the performance of the modified training procedure to the original procedure for PubTables-1M.
We use a validation set to select the best model, which for the new training procedure selects the model after 29 epochs.
Using the modified training procedure improves performance over the original, which establishes new baseline performance metrics.

In the rest of our experiments we train nine TATR models in total using nine different training datasets.
We train one model for each modified version of the FinTabNet dataset, a baseline model on PubTables-1M, and two additional models: one model for PubTables-1M combined with FinTabNet.a1 and one for PubTables-1M combined with FinTabNet.a6.
Each model is trained for 30 epochs, where an epoch is defined as 720,000 samples.
We evaluate each model's checkpoint after every epoch on a validation set from the same distribution as its training data as well as a separate validation set from ICDAR-2013.
We average the values of $\text{GriTS}_\text{Con}$ for the two validation sets and select the saved checkpoint yielding the highest score.

\subsection{FinTabNet self-evaluation}

\begin{figure*}[t]
  \centering
  \begin{subfigure}[b]{0.49\linewidth}
	\centering
	\includegraphics[width=6cm]{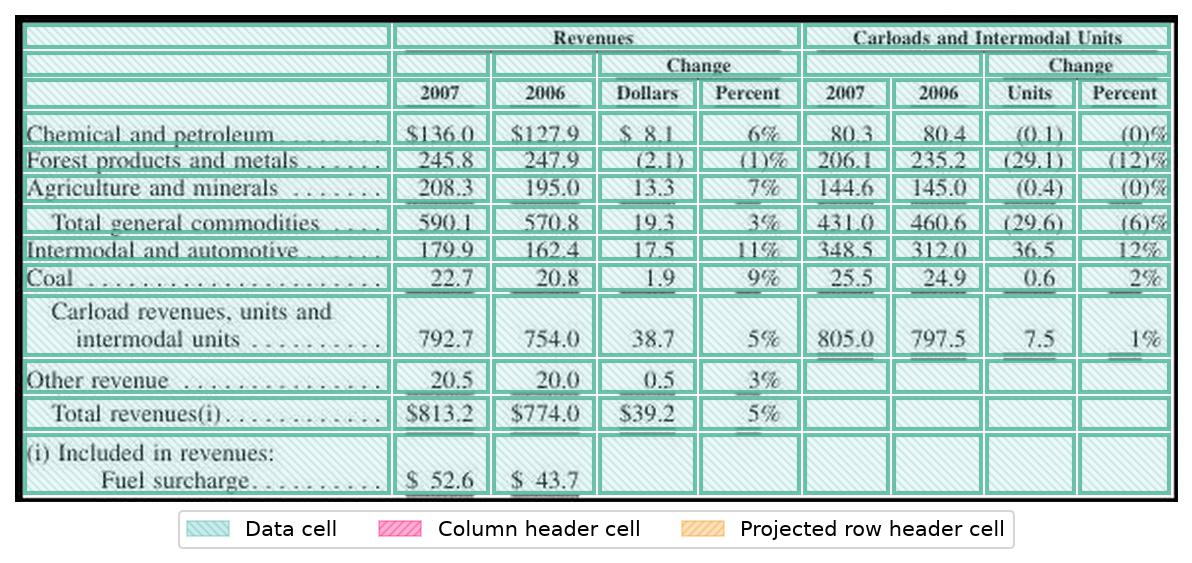}
    \caption{FinTabNet.a1}
    \label{subfig:overlap}
  \end{subfigure}
  \begin{subfigure}[b]{0.49\linewidth}
	\centering
	\includegraphics[width=6cm]{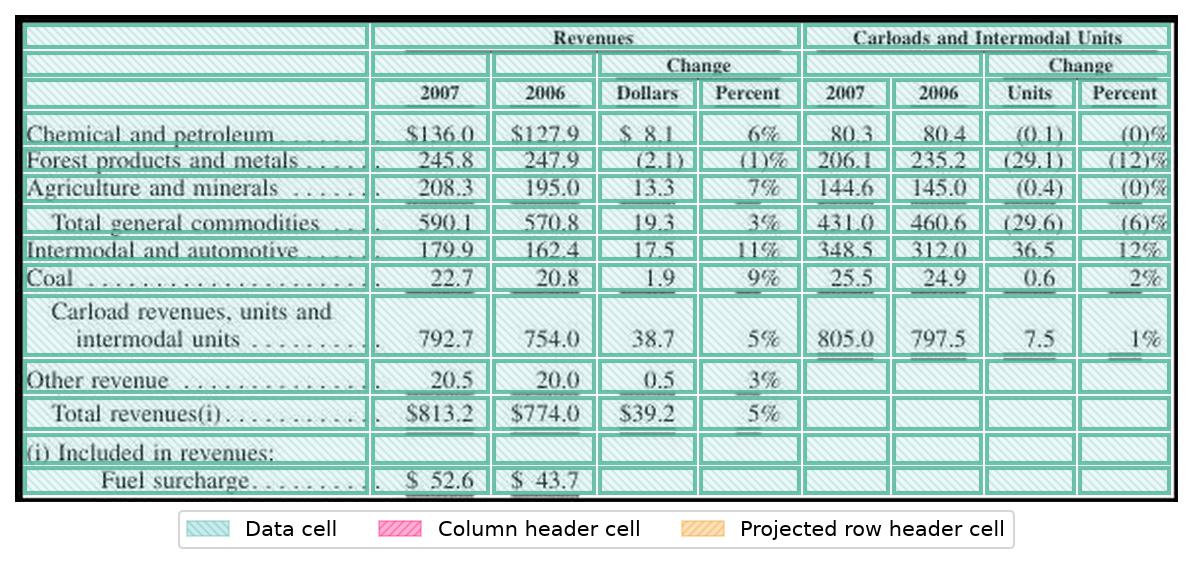}
    \caption{FinTabNet.a3}
    \label{subfig:without_overlap}
  \end{subfigure}
  \begin{subfigure}[b]{0.49\linewidth}
	\centering
	\includegraphics[width=6cm]{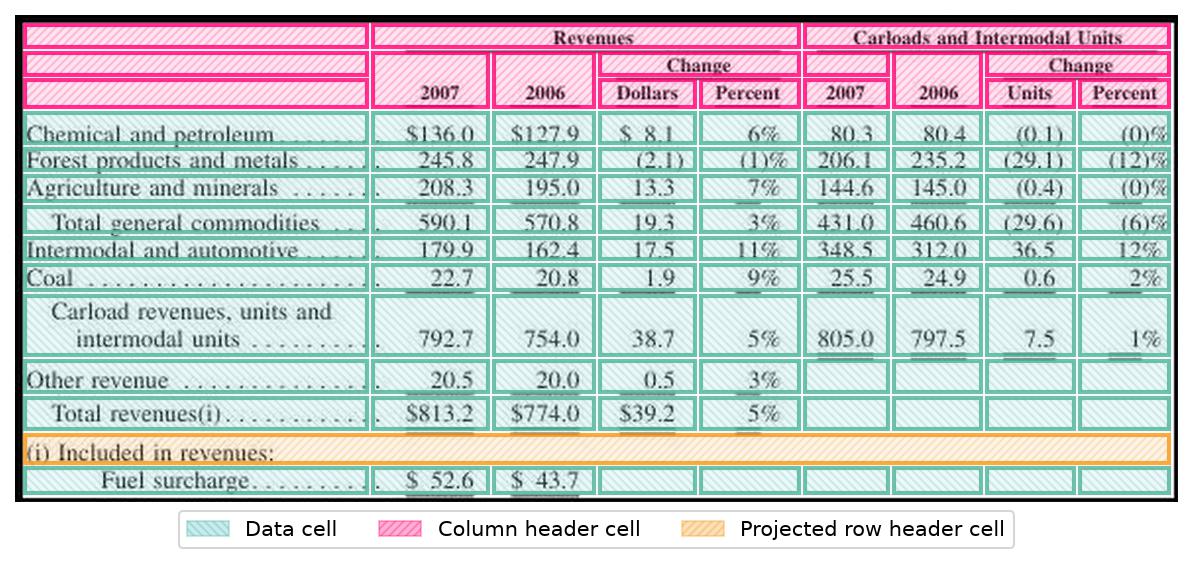}
    \caption{FinTabNet.a4}
    \label{subfig:without_overlap}
  \end{subfigure}
  \begin{subfigure}[b]{0.49\linewidth}
	\centering
	\includegraphics[width=6cm]{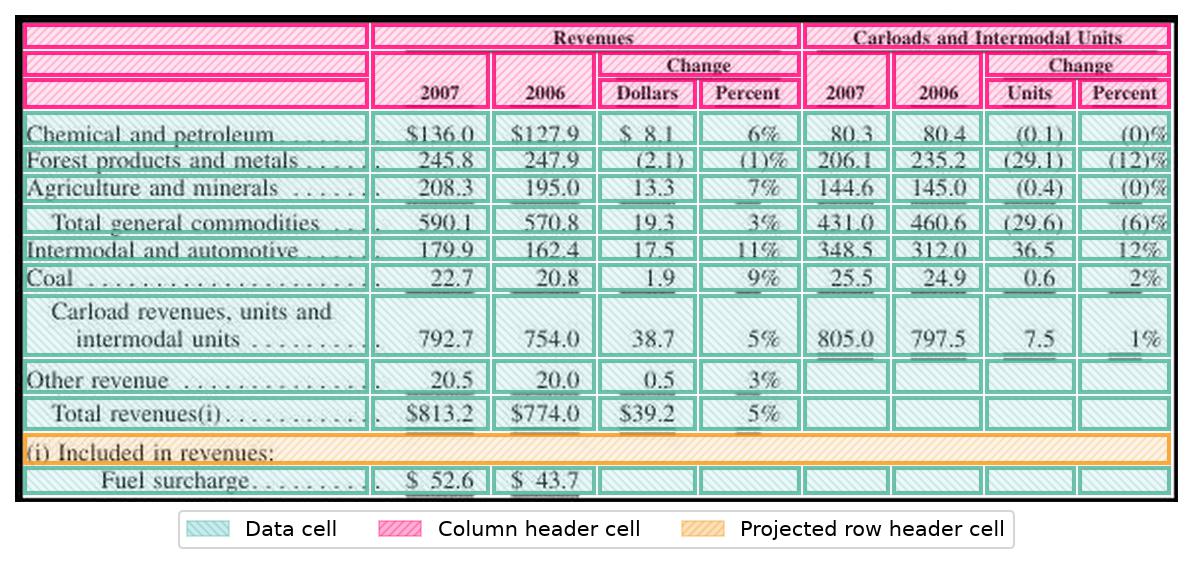}
    \caption{FinTabNet.a5}
    \label{subfig:without_overlap}
  \end{subfigure}
  \caption{Models trained on different versions of FinTabNet produce different table structure output for the same table from FinTabNet's test set. The differences include: how the last two rows are parsed, where the boundary between the first two columns occurs, and how the header is parsed into cells. (Column and row header annotations are also incorporated into the training starting with FinTabNet.a4).}
  \label{fig:fintabnet_qual}
\end{figure*}

\begin{table*}[!t]
  \scriptsize
  \caption{Object detection and TSR performance metrics for models trained and evaluated on each FinTabNet ablation's training and test sets. For model selection, we train each model for 30 epochs, evaluate the model on its validation set after every epoch, and choose the model with the highest $\text{Acc}_\text{Con}$ on the validation set.}
  \label{tab:fintabnet_self_test}
  \centering
  \begin{tabular}{lcrrrrrrrr}
    \toprule
    \textbf{Training Data} & \textbf{Epoch} & \textbf{AP} & $\textbf{AP}_\textbf{50}$ & $\textbf{AP}_\textbf{75}$ & \textbf{AR} & $\textbf{GriTS}_\textbf{Con}$ & $\textbf{GriTS}_\textbf{Loc}$ & $\textbf{GriTS}_\textbf{Top}$ & $\textbf{Acc}_\textbf{Con}$ \\
    \midrule
    FinTabNet.a1 & 27 & 0.867 & 0.972 & 0.932 & 0.910 & 0.9796 & 0.9701 & 0.9874 & 0.787 \\
    FinTabNet.a2 & 22 & \textbf{0.876} & 0.974 & 0.941 & 0.916 & 0.9800 & 0.9709 & 0.9874 & 0.768 \\
    FinTabNet.a3 & 14 & 0.871 & 0.975 & 0.942 & 0.910 & 0.9845 & 0.9764 & 0.9893 & 0.799\\
    FinTabNet.a4 & 30 & 0.874 & 0.974 & 0.946 & 0.919 & 0.9841 & 0.9772 & 0.9884 & 0.793 \\
    FinTabNet.a5 & 24 & 0.872 & \textbf{0.977} & \textbf{0.950} & 0.917 & \textbf{0.9861} & 0.9795 & \textbf{0.9897} & \textbf{0.828} \\
    FinTabNet.a6 & 26 & 0.875 & 0.976 & 0.948 & \textbf{0.921} & 0.9854 & \textbf{0.9796} & 0.9891 & 0.812 \\
    \midrule
  \end{tabular}
\end{table*}

In \cref{tab:fintabnet_self_test}, we present the results of each FinTabNet model evaluated on its own test set.
As can be seen, the complete set of dataset processing steps leads to an increase in $\textrm{Acc}_\textrm{Con}$ from 0.787 to 0.812.
In \cref{fig:fintabnet_qual}, we illustrate on a sample from the FinTabNet test set how the consistency of the output improves from TATR trained on FinTabNet.a1 to TATR trained on FinTabNet.a5.
Note that models trained on FinTabNet.a4 onward also learn header information in addition to structure information.

Models trained on FinTabNet.a5 and FinTabNet.a6 both have higher $\textrm{Acc}_\textrm{Con}$ than models trained on FinTabNet.a1 through FinTabNet.a3, despite the tables in a5 and a6 being more complex (and thus more difficult) overall than a1-a3 according to \cref{tab:stats}.
This strongly indicates that the results are driven primarily by increasing the consistency and cleanliness of the data, and not by reducing their inherent complexity.
Something else to note is that while $\text{GriTS}_\text{Loc}$, $\text{GriTS}_\text{Con}$, and $\text{Acc}_\text{Con}$ all increase as a result of the improvements to the data, $\text{GriTS}_\text{Top}$ is little changed.
$\text{GriTS}_\text{Top}$ measures only how well the model infers the table's cell layout alone, without considering how well the model locates the cells or extracts their text content.
This is strong evidence that the improvement in performance of the models trained on the modified data is driven primarily by there being more consistency in the annotations rather than the examples becoming less challenging.
However, this evidence is even stronger in the next section, when we evaluate the FinTabNet models on the modified ICDAR-2013 datasets instead.

\subsection{ICDAR-2013 evaluation}

\begin{figure*}[t]
  \centering
  \begin{subfigure}[b]{0.49\linewidth}
	\centering
	\includegraphics[width=6cm]{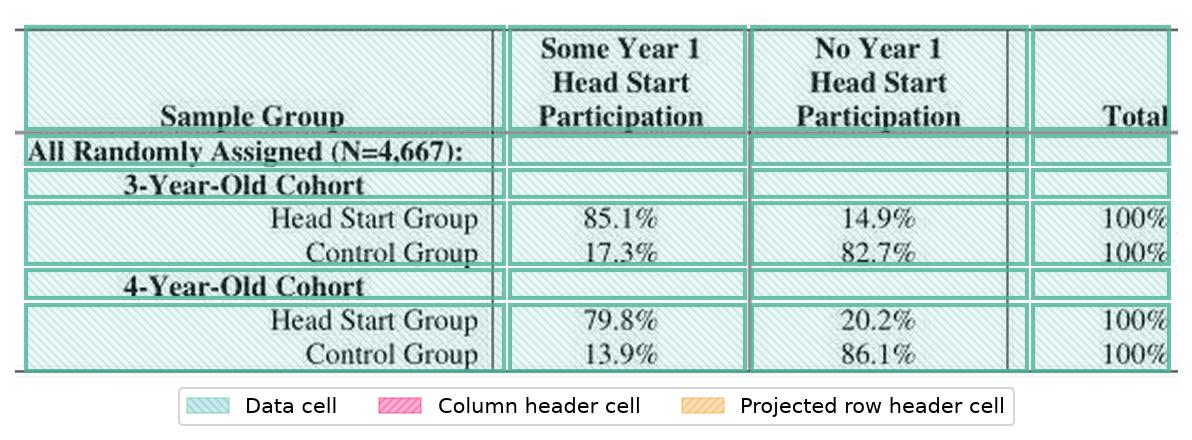}
    \caption{FinTabNet.a1}
    \label{subfig:overlap}
  \end{subfigure}
  \begin{subfigure}[b]{0.49\linewidth}
	\centering
	\includegraphics[width=6cm]{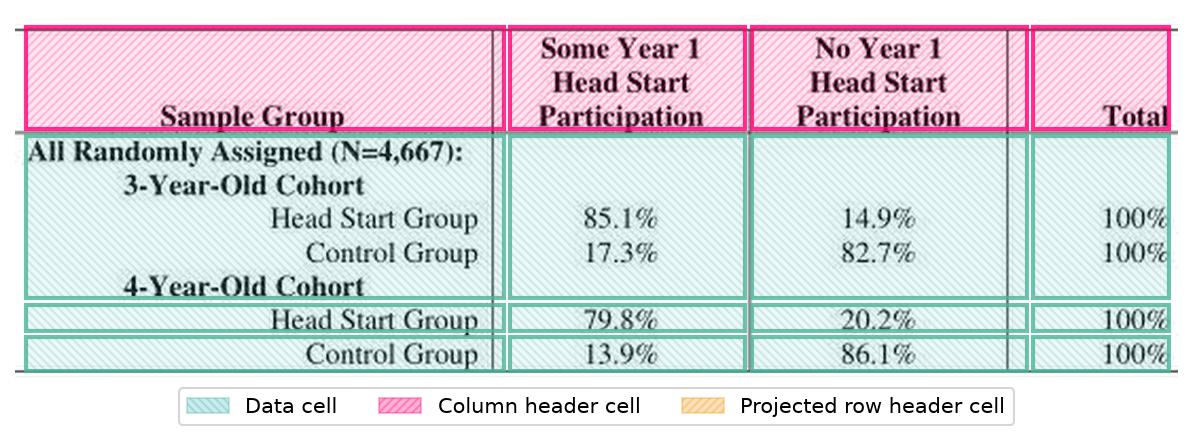}
    \caption{PubTables-1M}
    \label{subfig:without_overlap}
  \end{subfigure}
  \begin{subfigure}[b]{0.49\linewidth}
	\centering
	\includegraphics[width=6cm]{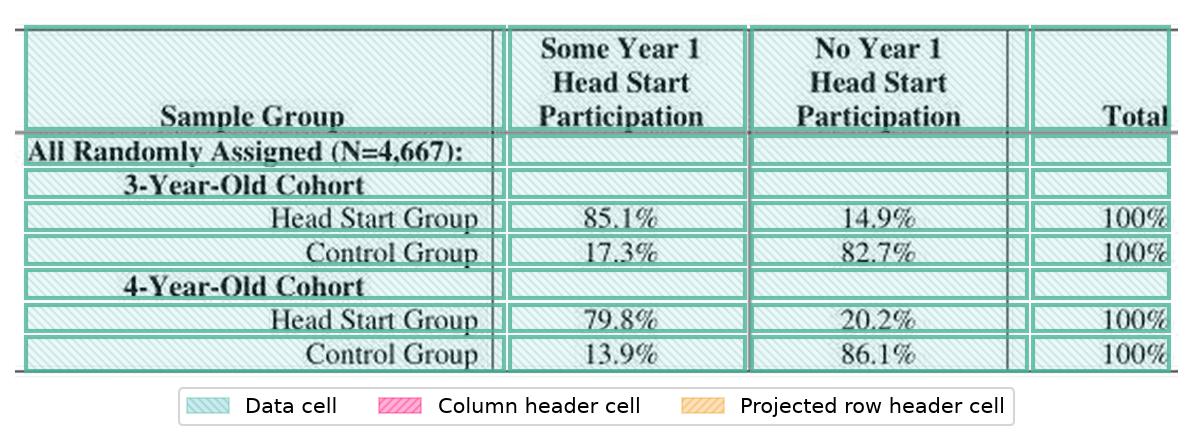}
    \caption{FinTabNet.a3}
    \label{subfig:without_overlap}
  \end{subfigure}
  \begin{subfigure}[b]{0.49\linewidth}
	\centering
	\includegraphics[width=6cm]{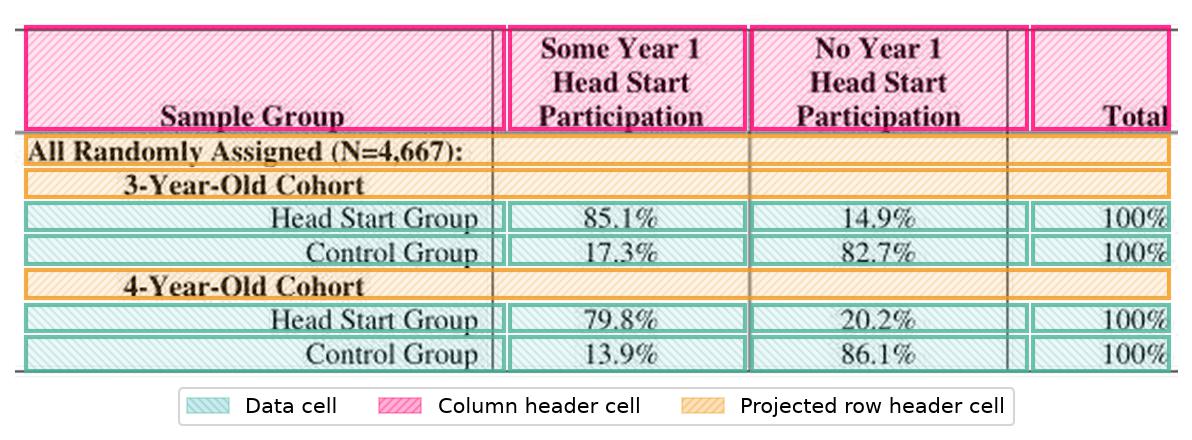}
    \caption{FinTabNet.a6}
    \label{subfig:without_overlap}
  \end{subfigure}
  \begin{subfigure}[b]{0.49\linewidth}
	\centering
	\includegraphics[width=6cm]{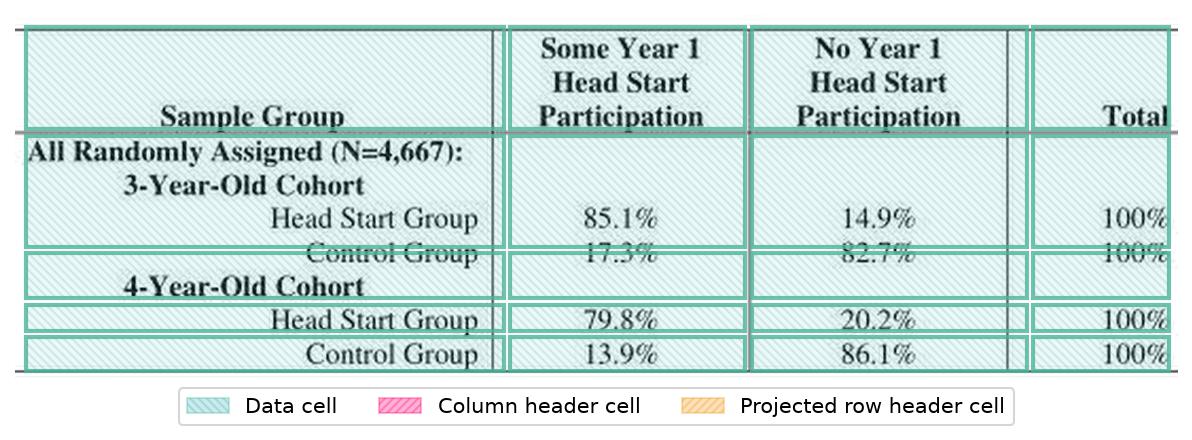}
    \caption{FinTabNet.a1 + PubTables-1M}
    \label{subfig:without_overlap}
  \end{subfigure}
  \begin{subfigure}[b]{0.49\linewidth}
	\centering
	\includegraphics[width=6cm]{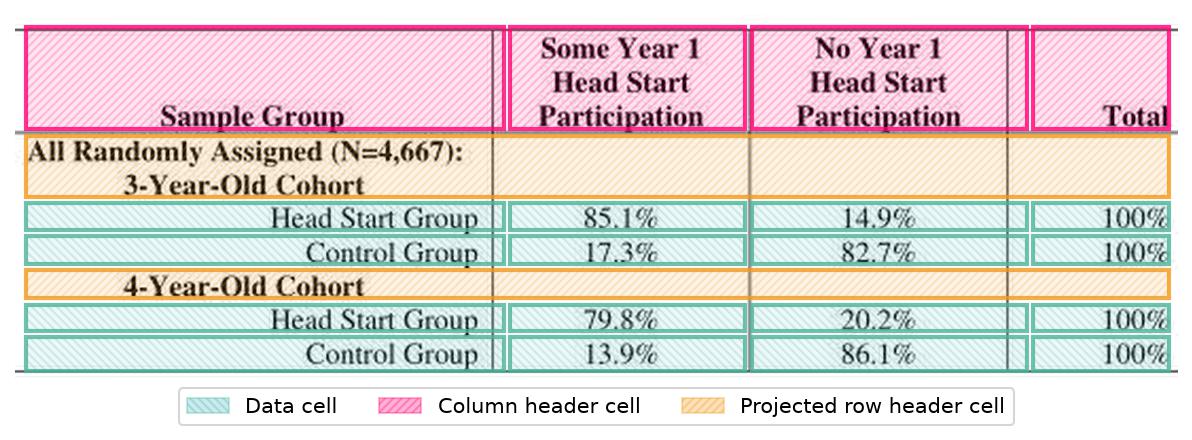}
    \caption{FinTabNet.a6 + PubTables-1M}
    \label{subfig:without_overlap}
  \end{subfigure}
  \caption{Table structure output for a table from the ICDAR-2013 test set for models trained on different datasets. This example highlights how both the variety of the examples in the training sets and the consistency with which the examples are annotated affect the result.}
  \label{fig:icdar2013_qual}
\end{figure*}

\begin{table*}[!t]
  \scriptsize
  \caption{Object detection and TSR performance metrics for models trained on each FinTabNet ablation's training set and evaluated on each ICDAR-2013 ablation.}
  \label{tab:icdar_test_results}
  \centering
  \begin{tabular}{lllrrrrrrrrr}
    \toprule
    \textbf{Training} & \textbf{Ep.} & \textbf{Test Data} & \textbf{AP} & $\textbf{AP}_\textbf{50}$ & $\textbf{AP}_\textbf{75}$ & $\textbf{DAR}_\textbf{C}$ & $\textbf{GriTS}_\textbf{C}$ & $\textbf{GriTS}_\textbf{L}$ & $\textbf{GriTS}_\textbf{T}$ & $\textbf{Acc}_\textbf{C}$\\
    \midrule
    FinTabNet.a1 & 28 & IC13.a1 & 0.670 & 0.859 & 0.722 & 0.8922 & 0.9390 & 0.9148 & 0.9503 & 0.417 \\
       & 28 & IC13.a2 & 0.670 & 0.859 & 0.723 & 0.9010 & 0.9384 & 0.9145 & 0.9507 & 0.411 \\
       & 20 & IC13.a3 & 0.445 & 0.573 & 0.471 & 0.8987 & 0.9174 & 0.8884 & 0.9320 & 0.411 \\
    \midrule
    FinTabNet.a2 & 27 & IC13.a1 & 0.716 & 0.856 & 0.778 & 0.9107 & 0.9457 & 0.9143 & 0.9536 & 0.436 \\
       & 27 & IC13.a2 & 0.714 & 0.856 & 0.778 & 0.9049 & 0.9422 & 0.9105 & 0.9512 & 0.430 \\
       & 29 & IC13.a3 & 0.477 & 0.576 & 0.516 & 0.8862 & 0.9196 & 0.8874 & 0.9336 & 0.405 \\
    \midrule
    FinTabNet.a3 & 25 & IC13.a1 & 0.710 & 0.851 & 0.765 & 0.9130 & 0.9462 & 0.9181 & 0.9546 & 0.462 \\
       & 25 & IC13.a2 & 0.710 & 0.850 & 0.764 & 0.9091 & 0.9443 & 0.9166 & 0.9538 & 0.456 \\
       & 28 & IC13.a3 & 0.470 & 0.571 & 0.505 & 0.8889 & 0.9229 & 0.8930 & 0.9346 & 0.418 \\
    \midrule
    FinTabNet.a4 & 30 & IC13.a1 & 0.763 & 0.935 & 0.817 & 0.9134 & 0.9427 & 0.9170 & 0.9516 & 0.551 \\
                     & 30 & IC13.a2 & 0.763 & 0.935 & 0.818 & 0.9088 & 0.9409 & 0.9155 & 0.9503 & 0.544 \\
                     & 25 & IC13.a3 & 0.765 & 0.944 & 0.832 & 0.9287 & 0.9608 & 0.9409 & 0.9703 & 0.589 \\
    \midrule
    FinTabNet.a5 & 30 & IC13.a1 & 0.774 & 0.939 & 0.838 & 0.9119 & 0.9412 & 0.9075 & 0.9456 & 0.494 \\
                     & 30 & IC13.a2 & 0.774 & 0.940 & 0.840 & 0.9098 & 0.9400 & 0.9057 & 0.9443 & 0.487 \\
                     & 16 & IC13.a3 & 0.773 & 0.956 & 0.854 & 0.9198 & 0.9548 & 0.9290 & 0.9624 & 0.551 \\
    \midrule
    FinTabNet.a6 & 23 & IC13.a1 & 0.760 & 0.927 & 0.830 & 0.9057 & 0.9369 & 0.9117 & 0.9497 & 0.500 \\
                     & 21 & IC13.a2 & 0.757 & 0.926 & 0.818 & 0.9049 & 0.9374 & 0.9106 & 0.9491 & 0.506 \\
                     & 20 & IC13.a3 & 0.757 & 0.941 & 0.840 & 0.9336 & 0.9625 & 0.9431 & 0.9702 & 0.646 \\
    \midrule
    \midrule
    PubTables-1M & 29 & IC13.a1 & 0.873 & 0.972 & 0.943 & 0.9440 & 0.9590 & 0.9462 & 0.9623 & 0.647 \\
                 & 29 & IC13.a2 & 0.873 & 0.972 & 0.941 & 0.9392 & 0.9570 & 0.9448 & 0.9608 & 0.639 \\
                 & 30 & IC13.a3 & 0.828 & 0.973 & 0.934 & 0.9570 & 0.9756 & 0.9700 & 0.9786 & 0.753 \\
    \midrule
    PubTables-1M + & 25 & IC13.a1 & 0.872 & 0.970 & 0.940 & 0.9543 & 0.9678 & 0.9564 & 0.9720 & 0.686 \\
    FinTabNet.a1 & 25 & IC13.a2 & 0.871 & 0.970 & 0.939 & 0.9501 & 0.9655 & 0.9543 & 0.9704 & 0.677 \\
                 & 28 & IC13.a3 & 0.820 & 0.949 & 0.911 & 0.9630 & 0.9787 & 0.9702 & 0.9829 & 0.785 \\
    \midrule
    PubTables-1M + & 28 & IC13.a1 & 0.881 & 0.977 & 0.953 & 0.9687 & 0.9678 & 0.9569 & 0.9705 & 0.679 \\
    FinTabNet.a6 & 28 & IC13.a2 & 0.880 & 0.975 & 0.951 & 0.9605 & 0.9669 & 0.9566 & 0.9702 & 0.671 \\
                 & 29 & IC13.a3 & 0.826 & 0.974 & 0.934 & 0.9648 & 0.9811 & 0.9750 & 0.9842 & 0.810 \\
    \midrule
  \end{tabular}
\end{table*}

In the next set of results, we evaluate all nine models on all three versions of the ICDAR-2013 dataset.
These results are intended to show the combined effects that improvements to both the training data and evaluation data have on measured performance.
Detailed results are given in \cref{tab:icdar_test_results} and a visualization of the results for just $\text{Acc}_\text{Con}$ is given in \cref{fig:summary}.

Note that all of the improvements to the ICDAR-2013 annotations are verified manually and no tables are removed from the dataset.
So observed improvements to performance from ICDAR-2013.a1 to ICDAR-2013.a3 can only be due to cleaner and more consistent evaluation data.
This improvement can be observed clearly for each trained model from FinTabNet.a4 onward.
Improvements to ICDAR-2013 alone are responsible for measured performance for the model trained on PubTables-1M and FinTabNet.a6 combined increasing significantly, from 68\% to 81\% $\text{Acc}_\text{Con}$.

Evaluating improvements to FinTabNet on ICDAR-2013.a3, we also observe a substantial increase in performance, from 41.1\% to 64.6\% $\text{Acc}_\text{Con}$.
This indicates that not only do improvements to FinTabNet improve its own self-consistency, but also significantly improve its performance on challenging real-world test data.
The improvement in performance on test data from both FinTabNet and ICDAR-2013 clearly indicates that there is a significant increase in the consistency of the data across both benchmark datasets.

These results also hold for FinTabNet combined with PubTables-1M.
The final model trained with PubTables-1M and FinTabNet.a6 combined outperforms all other baselines, achieving a final $\text{Acc}_\text{Con}$ of 81\%.
Few prior works report this metric on ICDAR-2013 or any benchmark dataset for TSR.
As we discuss previously, this is likely due to the fact that measured accuracy depends not only on model performance but the quality of the ground truth annotation.
These results simultaneously establish new performance baselines and indicate that the evaluation data is clean enough for this metric to be a useful measure of model performance.

While TATR trained on PubTables-1M and FinTabNet.a6 jointly performs best overall, in \cref{fig:fintabnet_qual} we highlight an interesting example from the ICDAR-2013 test set where TATR trained on FinTabNet.a6 alone performs better.
For this test case, TATR trained on the baseline FinTabNet.a1 does not fully recognize the table's structure, while TATR trained on the FinTabNet.a6 dataset does.
Surprisingly, adding PubTables-1M to the training makes recognition for this example worse.
We believe cases like this suggest the potential for future work to explore additional ways to leverage PubTables-1M and FinTabNet.a6 to improve joint performance even further.

\begin{figure*}[t]
	\centering
	\resizebox{\linewidth}{!}{
	\begin{tikzpicture}[scale=0.9]
		\definecolor{light gray}{RGB}{220, 220, 220}
		\begin{axis}[height=7cm,width=18cm,grid, grid style={light gray, line cap=round}, xmin=0.5, xmax=30.5, ymin=0.935, ymax=0.986, legend pos=south east, legend columns=3,
				   xlabel={Epoch},title={ICDAR-2013.a3 Model Selection}, ylabel={$\text{GriTS}_\text{Con}$}]
		\addplot [blue, thick, line cap=round, mark=square, dotted, mark options={solid}, mark size=3pt]
			table {FinTabNet.a1_ICDAR2013_model_selection.txt};
		\addlegendentry[]{FinTabNet.a1}
		\addplot [red, line cap=round, mark=star, dotted, mark size=3pt, mark options={solid}]
			table {FinTabNet.a2_ICDAR2013_model_selection.txt};
		\addlegendentry[]{FinTabNet.a2}
		\addplot [orange, line cap=round, mark=diamond, dotted, mark size=3pt, mark options={solid}]
			table {FinTabNet.a3_ICDAR2013_model_selection.txt};
		\addlegendentry[]{FinTabNet.a3}
		\addplot [olive, thick, line cap=round, mark=x, dotted, mark options={solid},
		mark size=3pt]
			table {FinTabNet.a4_ICDAR2013_model_selection.txt};
		\addlegendentry[]{FinTabNet.a4}
		\addplot [teal, thick, line cap=round, mark=+, dotted, mark options={solid},
		mark size=3pt]
			table {FinTabNet.a5_ICDAR2013_model_selection.txt};
		\addlegendentry[]{FinTabNet.a5}
		\addplot [magenta, thick, line cap=round, mark=pentagon, dotted, mark options={solid},
		mark size=3pt]
			table {FinTabNet.a6_ICDAR2013_model_selection.txt};
		\addlegendentry[]{FinTabNet.a6}
		\addplot [cyan, thick, line cap=round, mark=10-pointed star, dotted, mark options={solid},
		mark size=3pt]
			table {PubTables-1M_ICDAR2013_model_selection.txt};
		\addlegendentry[]{PubTables-1M}
		\addplot [pink, thick, line cap=round, mark=Mercedes star flipped, dotted, mark options={solid},
		mark size=3pt]
			table {PubTables-1M_FinTabNet.a1_ICDAR2013_model_selection.txt};
		\addlegendentry[]{PT1M + FTN.a1}
		\addplot [brown, thick, line cap=round, mark=asterisk, dotted, mark options={solid},
		mark size=3pt]
			table {PubTables-1M_FinTabNet.a6_ICDAR2013_model_selection.txt};
		\addlegendentry[]{PT1M + FTN.a6}
		\addplot[color=blue, mark=o, mark options={line width=3, opacity=0.6}, mark size=7pt]
			coordinates {
			      (20, 0.9605)};
		\addplot[solid, fill=red, mark=o, mark options={solid, draw=red, fill=red, line width=3, opacity=0.6}, mark size=7pt]
			coordinates {
			      (29, 0.9629)};
		\addplot[solid, mark=o, mark options={solid, draw=orange, fill=orange, line width=3, opacity=0.6}, mark size=7pt]
			coordinates {
			      (28, 0.9652)};
		\addplot[color=olive, mark=o, mark options={line width=3pt, opacity=0.6}, mark size=7pt]
			coordinates {
			      (25, 0.9711)};
		\addplot[color=teal, mark=o, mark options={line width=3pt, opacity=0.6}, mark size=7pt]
			coordinates {
			      (16, 0.9733)};
		\addplot[color=magenta, mark=o, mark options={line width=3pt, opacity=0.6}, mark size=7pt]
			coordinates {
			      (20, 0.9698)};
		\addplot[color=cyan, mark=o, mark options={line width=3pt, opacity=0.6}, mark size=7pt]
			coordinates {
			      (30, 0.9686)};
		\addplot[color=pink, mark=o, mark options={line width=3pt, opacity=0.8}, mark size=7pt]
			coordinates {
			      (28, 0.9798)};
		\addplot[color=brown, mark=o, mark options={line width=3pt, opacity=0.6}, mark size=7pt]
			coordinates {
			      (29, 0.9839)};
		\end{axis}
	\end{tikzpicture}
	}
    \label{fig:plot}
	\caption{The performance of each model after every epoch of training evaluated on its validation set. The training epoch with the best performance for each model is circled. PT1M = PubTables-1M, FTN = FinTabNet.}
	\label{fig:model_selection}
\end{figure*}

\begin{figure*}[t]
	\centering
	\resizebox{\linewidth}{!}{
	\begin{tikzpicture}[scale=0.9]
		\definecolor{light gray}{RGB}{220, 220, 220}
		\begin{axis}[height=8cm,width=18cm,grid, grid style={light gray, line cap=round}, xmin=0.5, xmax=30.5, ymin=0.2, ymax=0.83, legend pos=south east, legend columns=3,
				   xlabel={Epoch},title={ICDAR-2013.a3 evaluation}, ylabel={$\text{Acc}_\text{Con}$}]
		\addplot [blue, thick, line cap=round, mark=square, dotted, mark options={solid}, mark size=3pt]
			table {FinTabNet.a1.txt};
		\addlegendentry[]{FinTabNet.a1}
		\addplot [red, line cap=round, mark=star, dotted, mark size=3pt, mark options={solid}]
			table {FinTabNet.a2.txt};
		\addlegendentry[]{FinTabNet.a2}
		\addplot [orange, line cap=round, mark=diamond, dotted, mark size=3pt, mark options={solid}]
			table {FinTabNet.a3.txt};
		\addlegendentry[]{FinTabNet.a3}
		\addplot [olive, thick, line cap=round, mark=x, dotted, mark options={solid},
		mark size=3pt]
			table {FinTabNet.a4.txt};
		\addlegendentry[]{FinTabNet.a4}
		\addplot [teal, thick, line cap=round, mark=+, dotted, mark options={solid},
		mark size=3pt]
			table {FinTabNet.a5.txt};
		\addlegendentry[]{FinTabNet.a5}
		\addplot [magenta, thick, line cap=round, mark=pentagon, dotted, mark options={solid},
		mark size=3pt]
			table {FinTabNet.a6.txt};
		\addlegendentry[]{FinTabNet.a6}
		\addplot [cyan, thick, line cap=round, mark=10-pointed star, dotted, mark options={solid},
		mark size=3pt]
			table {PubTables-1M.txt};
		\addlegendentry[]{PubTables-1M}
		\addplot [pink, thick, line cap=round, mark=Mercedes star flipped, dotted, mark options={solid},
		mark size=3pt]
			table {PubTables-1M_FinTabNet.a1.txt};
		\addlegendentry[]{PT1M + FTN.av1}
		\addplot [brown, thick, line cap=round, mark=asterisk, dotted, mark options={solid},
		mark size=3pt]
			table {PubTables-1M_FinTabNet.a6.txt};
		\addlegendentry[]{PT1M + FTN.av6}
		\addplot[color=blue, mark=o, mark options={line width=3, opacity=0.6}, mark size=7pt]
			coordinates {
			      (20, 0.411)};
		\addplot[solid, fill=red, mark=o, mark options={solid, draw=red, fill=red, line width=3, opacity=0.6}, mark size=7pt]
			coordinates {
			      (29, 0.405)};
		\addplot[solid, mark=o, mark options={solid, draw=orange, fill=orange, line width=3, opacity=0.6}, mark size=7pt]
			coordinates {
			      (28, 0.418)};
		\addplot[color=olive, mark=o, mark options={line width=3pt, opacity=0.6}, mark size=7pt]
			coordinates {
			      (25, 0.589)};
		\addplot[color=teal, mark=o, mark options={line width=3pt, opacity=0.6}, mark size=7pt]
			coordinates {
			      (16, 0.551)};
		\addplot[color=magenta, mark=o, mark options={line width=3pt, opacity=0.6}, mark size=7pt]
			coordinates {
			      (20, 0.646)};
		\addplot[color=cyan, mark=o, mark options={line width=3pt, opacity=0.6}, mark size=7pt]
			coordinates {
			      (30, 0.753)};
		\addplot[color=pink, mark=o, mark options={line width=3pt, opacity=0.8}, mark size=7pt]
			coordinates {
			      (28, 0.785)};
		\addplot[color=brown, mark=o, mark options={line width=3pt, opacity=0.6}, mark size=7pt]
			coordinates {
			      (29, 0.810)};
		\end{axis}
	\end{tikzpicture}
	}
    \label{fig:plot}
	\caption{The performance of each model after every epoch of training evaluated on ICDAR-2013.a3. The training epoch with the best performance for each model on its own validation set is circled. PT1M = PubTables-1M, FTN = FinTabNet.}
	\label{fig:metrics}
\end{figure*}

\section{Limitations}

While we demonstrated clearly that removing annotation inconsistencies between TSR datasets improves model performance, we did not directly measure the accuracy of the automated alignment procedure itself.
Instead, we minimized annotation mistakes by introducing quality control checks and filtering out any tables in FinTabNet whose annotations failed these tests.
But it is possible that the alignment procedure may introduce its own mistakes, some of which may not be caught by the quality control checks.
It is also possible that some oversegmented tables could require domain knowledge specific to the table content itself to infer their canonicalized structure, which could not easily be incorporated into an automated canonicalization algorithm.
Finally, the quality control checks themselves may in some cases be too restrictive, filtering out tables whose annotations are actually correct.
Therefore it is an open question to what extent model performance could be improved by further improving the alignment between these datasets or by improving the quality control checks.

\section{Conclusion}

In this work we addressed the problem of \emph{misalignment} in benchmark datasets for table structure recognition.
We adopted the Table Transformer (TATR) model and three standard benchmarks for TSR---FinTabNet, PubTables-1M, and ICDAR-2013---and removed significant errors and inconsistencies between them.
After data improvements, performance of TATR on ICDAR-2013 increased substantially from 42\% to 65\% exact match accuracy when trained on FinTabNet and 65\% to 75\% when trained on PubTables-1M.
In addition, we trained TATR on the final FinTabNet and PubTables-1M datasets combined, establishing new improved baselines of 0.965 DAR and 81\% exact match accuracy on the ICDAR-2013 benchmark through data improvements alone.
Finally, we demonstrated through ablations that canonicalization has a significantly positive effect on the performance improvements across all three datasets.

\section{Acknowledgments}
We would like to thank the anonymous reviewers for helpful feedback while preparing this manuscript.

\bibliographystyle{splncs04}  
\bibliography{references}

\end{document}